\documentclass[sigconf,nonacm]{acmart}

\usepackage{booktabs} %
\usepackage{graphicx}
\usepackage{amsmath}
\usepackage{booktabs}
\usepackage{multirow}
\usepackage{colortbl}
\usepackage{makecell}
\usepackage{diagbox}

\usepackage{lipsum}

\newcommand\blfootnote[1]{%
  \begingroup
  \renewcommand\thefootnote{}\footnote{#1}%
  \addtocounter{footnote}{-1}%
  \endgroup
}

\usepackage{xcolor}
\definecolor{myred}{rgb}{0.8,0,0}
\definecolor{mygreen}{rgb}{0,0.8,0}

\usepackage{pifont}%
\newcommand{\cmark}{{\color{mygreen}\ding{51}}}
\newcommand{\xmark}{{\color{myred}\ding{55}}}

\usepackage[ruled]{algorithm2e} %

\SetAlFnt{\small}
\SetAlCapFnt{\small}
\SetAlCapNameFnt{\small}
\SetAlCapHSkip{0pt}

\AtBeginDocument{%
  \providecommand\BibTeX{{%
    \normalfont B\kern-0.5em{\scshape i\kern-0.25em b}\kern-0.8em\TeX}}}

\copyrightyear{2023}
\acmYear{2023}
\setcopyright{acmlicensed}
\acmConference[SA Technical Communications '23]{SIGGRAPH Asia 2023 Technical Communications}{December 12--15, 2023}{Sydney, NSW, Australia}
\acmBooktitle{SIGGRAPH Asia 2023 Technical Communications (SA Technical Communications '23), December 12--15, 2023, Sydney, NSW, Australia}
\acmPrice{15.00}
\acmDOI{10.1145/3610543.3626173}
\acmISBN{979-8-4007-0314-0/23/12}

\acmSubmissionID{141}

\begin{document}
\title{EasyVolcap: Accelerating Neural Volumetric Video Research}

\author{Zhen Xu}
\author{Tao Xie}
\author{Sida Peng}
\author{Haotong Lin}
\email{zhenx@zju.edu.cn}
\email{xbillowy@gmail.com}
\email{pengsida@zju.edu.cn}
\email{haotongl@outlook.com}
\affiliation{%
    \institution{Zhejiang University}
    \country{China}
}

\author{Qing Shuai}
\author{Zhiyuan Yu}
\author{Guangzhao He}
\email{s_q@zju.edu.cn}
\email{zyuaq@ust.hk}
\email{alexhe2000@zju.edu.cn}
\affiliation{%
    \institution{HKUST}
    \country{China}
}
\affiliation{%
    \institution{Zhejiang University}
    \country{China}
}

\author{Jiaming Sun}
\author{Hujun Bao}
\author{Xiaowei Zhou}
\authornote{Corresponding author.}
\email{suenjiaming@gmail.com}
\email{bao@cad.zju.edu.cn}
\email{xwzhou@zju.edu.cn}
\affiliation{%
    \institution{Image Derivative Inc.}
    \country{China}
}
\affiliation{%
    \institution{Zhejiang University}
    \country{China}
}

\renewcommand{\shortauthors}{Xu, Z. et al.}

% \begin{CCSXML}
%     <ccs2012>
%     <concept>
%     <concept_id>10010147.10010371.10010382.10010385</concept_id>
%     <concept_desc>Computing methodologies~Image-based rendering</concept_desc>
%     <concept_significance>500</concept_significance>
%     </concept>
%     </ccs2012>
% \end{CCSXML}
% \ccsdesc[500]{Computing methodologies~Image-based rendering}

% \keywords{Novel view synthesis, neural rendering, neural radiance field}

\newcommand{\easyvolcap}{\textcolor{black}{EasyVolcap}\xspace}

\begin{teaserfigure}
    \centering
    % \vspace*{-1.0em}
    \includegraphics[width=1.0\textwidth]{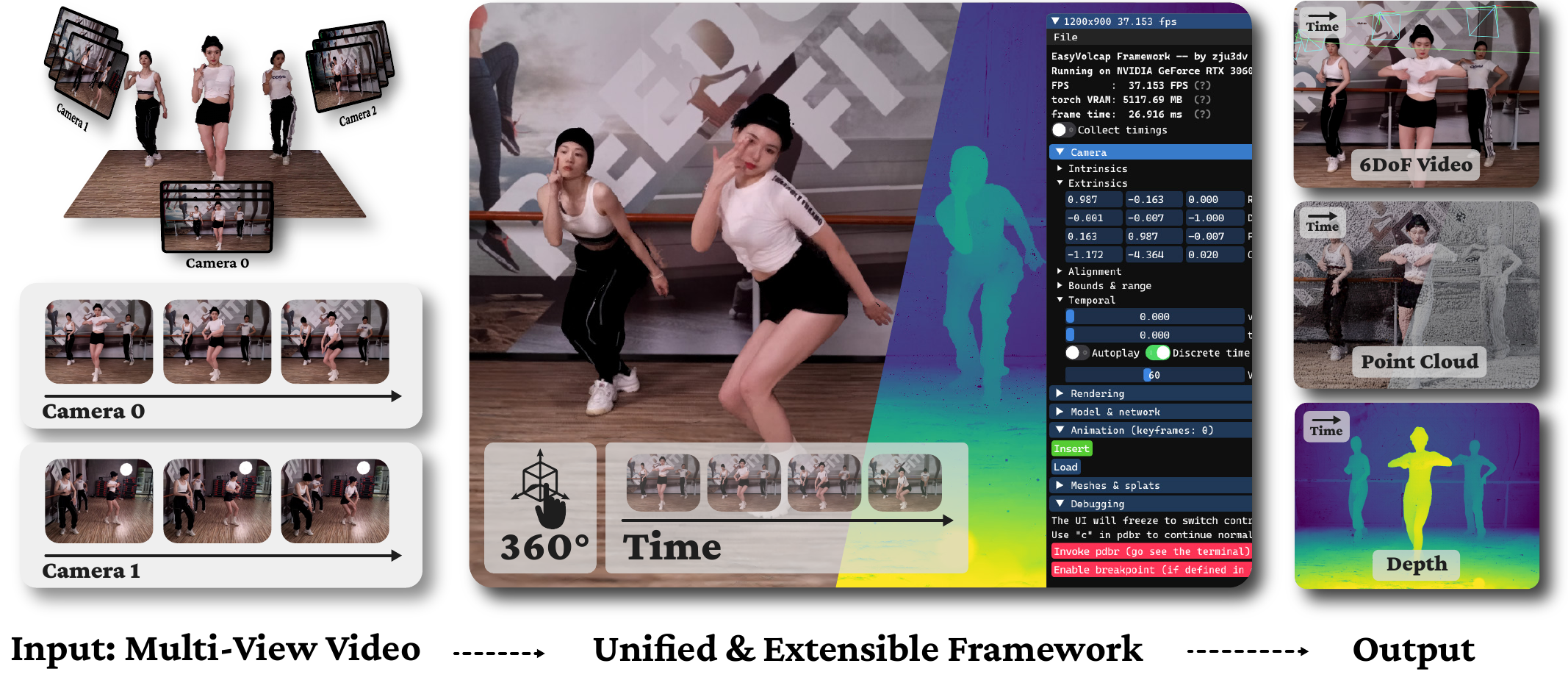}
    \vspace*{-2.0em}
    \caption{
        \textbf{\easyvolcap is a Python \& PyTorch library for accelerating volumetric video research, particularly in the area of neural dynamic scene representation, reconstruction, and rendering.} Given multi-view video input, \easyvolcap streamlines the process of data preprocessing, 4D reconstruction, and rendering of dynamic scenes.
        Our source code is available \href{https://github.com/zju3dv/EasyVolcap}{here at GitHub}.
    }
    \vspace*{1.5em}
    \label{fig:teaser}
\end{teaserfigure}
\maketitle

\section{Introduction}
Volumetric video is a technology that digitally records dynamic events such as artistic performances, sporting events, and remote conversations. When acquired, such volumography can be viewed from any viewpoint and timestamp on flat screens, 3D displays, or VR headsets, enabling immersive viewing experiences and more flexible content creation in a variety of applications such as sports broadcasting, video conferencing, gaming, and movie productions.
With the recent advances and fast-growing interest in neural scene representations for volumetric video,
there is an urgent need for a unified open-source library to streamline the process of volumetric video capturing, reconstruction, and rendering for both researchers and
non-professional users to develop various algorithms and applications of this emerging technology.
\blfootnote{Source code: \href{https://github.com/zju3dv/EasyVolcap}{https://github.com/zju3dv/EasyVolcap}}

\begin{figure*}
    \centering
    \vspace*{-1.0em}
    \includegraphics[width=1.0\textwidth]{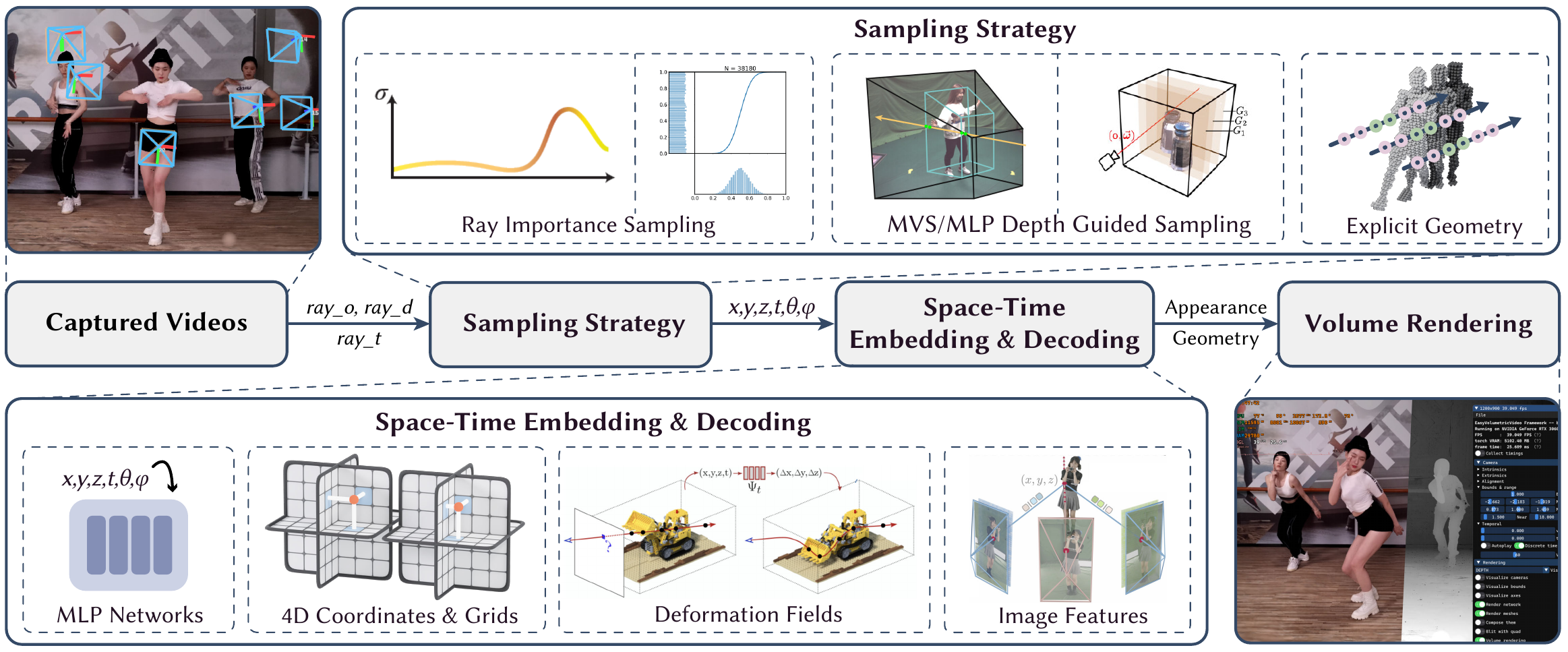}
    \vspace*{-2.0em}
    \caption{
        \textbf{Framework design of \easyvolcap.}
        Given captured videos, \easyvolcap samples space-time rays for rendering.
        Ray sampling strategies are applied to generate 4D point coordinates.
        These points are then encoded with space-time embedders, decoded for physical values (density, color, etc.), and then fed into the volume renderer to obtain the final rendering.
    }
    \label{fig:framework}
\end{figure*}

In this paper, we present \textit{\textbf{EasyVolcap}}, a Python \& Pytorch library for accelerating neural volumetric video research with the goal of unifying the process of multi-view data processing, 4D scene reconstruction, and efficient dynamic volumetric video rendering.
Given the insight into the most popular paradigm of dynamic scene reconstruction and rendering methods,
we build a well-structured unified pipeline for 4D scene reconstruction,
which is composed of a 4D-aware feature embedder and an MLP-based regressor as shown in Figure~\ref{fig:framework}.
Given the spacetime coordinates, the feature embedder maps them to a high-dimensional feature vector, which is then passed through the geometry and appearance MLPs to regress the final output density and color.
Moreover, we provide a set of easy-to-use tools for 4D volumetric video researchers like a native high-performance viewer marrying the extensibility of Python and the power of OpenGL and CUDA, and a robust training and evaluation loop for multi-view datasets.
Compared to NeRFStudio \cite{tancik2023nerfstudio}, which focuses on NeRF-based \cite{mildenhall2021nerf} static scene modeling,
\easyvolcap's data-loading procedure, unified pipeline, and other logistic systems are all specifically designed for dynamic 3D scenes as shown in Table~\ref{tab:comparison}.
Note that \easyvolcap trivially supports taking a static 3D dataset or a monocular dynamic dataset as input since they can be considered as special cases of the volumetric video with only one frame or only one camera.
We hope that by building a readily accessible open-source library on 4D volumetric video, future researchers on this topic could more easily express their creativity, develop brand-new algorithms, and discover groundbreaking insights.

\begin{table}[b]
    \centering
    \caption{
        \textbf{Comparison of different neural volumetric video frameworks.} NeRFStudio \cite{tancik2023nerfstudio} and SDFStudio \cite{Yu2022SDFStudio} are designed for static scenes and do not support multi-view data input.
        NerfAcc \cite{li2023nerfacc} and Kaolin-Wisp \cite{KaolinWispLibrary} do not support playback of dynamic volumetric content.
        Our framework supports both multi-view video datasets and playback of dynamic volumetric videos.
    }
    \resizebox{1.0\linewidth}{!}{
        \begin{tabular}{l*{3}{m{1.6cm}}}
            \toprule
            \textbf{Framework}                     & \textbf{Multi-View Video Dataset} & \textbf{Volumetric
 Video Player
  } \\
            \midrule
            NerfStudio \cite{tancik2023nerfstudio} & \xmark                            & \cmark                         \\
            SDFStudio\cite{Yu2022SDFStudio}        & \xmark                            & \xmark                         \\
            NerfAcc\cite{li2023nerfacc}            & \xmark                            & \xmark                         \\
            Kaolin-Wisp\cite{KaolinWispLibrary}    & \xmark                            & \xmark                         \\
            \midrule
            \easyvolcap                                   & \cmark                            & \cmark                         \\
            \bottomrule
        \end{tabular}
    }
    \label{tab:comparison}
\end{table}

\section{Framework Design}

We develop a unified framework after studying the recurring patterns of prominent 4D volumetric video methods where components can even be directly swapped from the command line.
This unified pipeline includes a 2D and 3D sampler, a set of space-time 4D embedding structures,
a deformation module to apply flow-like tracking on the underlying 3D structure, an appearance or transient embedding where appearance-only tunes could be applied,
and finally, a set of regressors to convert embedded features to the final output.
Figure~\ref{fig:framework} provides a detailed illustration of the framework's unified pipeline.
Moreover, when a novel algorithm requires a completely different network organization, it is also trivially easy to swap out the core components of the fixed pipeline.

\paragraph{Input \& dataset representation.} 
\easyvolcap provides a centralized but inheritable dataset management system where the most basic form input is an unstructured image tensor of $[n_{frame}, n_{view}]$. 
Numerous optimizations like dataset sharding for multi-gpu training, input data compression (which is never a problem until you add a ${time}$ dimension) with a custom addressable but unstructured tensor class, flexible input formats where pixel mask, pixel importance, and visual-hull-based bounding box determination are all a single switch away. 
One can also easily optimize the camera parameters thanks to the optimizable camera residual applied before sampling.

\paragraph{Point \& ray sampling strategy.} 
Typical NeRF-based \cite{mildenhall2021nerf} rendering requires the conversion from camera rays to points sampled on the ray before querying the network and performing volume rendering. 
The sampler family of \easyvolcap unifies this process with a corpus of point samplers like a uniform sampler, a disparity-based sampler \cite{barron2022mipnerf360}, a coarse-to-fine importance sampler \cite{mildenhall2021nerf}, a human shape guided \cite{peng2022animatable} sampler and a cost-volume-based depth-guided sampler \cite{lin2022efficient}. 
Optimizable parameters like the 2D feature extractors of cost-volume builder are also easily introduced in this stage.

\paragraph{Space-time feature embedding.} 
At the core of the main-stream volumetric representations is a set of 4D-aware encodings \cite{fridovich2023k}, either computed from representative implicit structures or fast explicit proxies. 
\easyvolcap embraces this concept by providing a KPlanes-style \cite{fridovich2023k} decomposed feature embedder and a composable 4D embedder (positional encoding \cite{mildenhall2021nerf}, multi-resolution hash encoding \cite{muller2022instant} and latent-codes \cite{peng2022animatable}).
The process of implementing and benchmarking a novel 4D representation is also streamlined thanks to \easyvolcap's robust registration and configuration system.

\paragraph{Deformation \& flow composition.} 
Several other prominent volumetric video research proposed components to model the deformation of the dynamic scene using scene flow \cite{wang2023tracking}, deformation fields \cite{park2021nerfies} or hyper-networks \cite{park2021hypernerf}. 
As an effective way to model the dynamic nature of the volumetric video, \easyvolcap also provides a set of deformation modules that can be easily applied along with their respective regularizations.

\paragraph{Appearance \& transient feature embedding.} 
\cite{zhang2020nerf++} propose to model the dynamic appearance of a time-varying scene with an appearance-specific latent embedding or tailored transient functions. 
Thanks to the modular nature of \easyvolcap's unified pipeline, such transient embedding is easily applied at the appearance embedding stage.

\paragraph{Output Regressor.} 
Given the embedded feature of a four-dimensional coordinate, \easyvolcap provides a set of physical property regressors to differentiably translate a neural descriptor in the 4D space to an actual output, such as volume density \cite{mildenhall2021nerf}, signed distance \cite{wang2021neus}, RGB color \cite{mildenhall2021nerf}, or Spherical Harmonics coefficients \cite{yu2021plenoctrees}.

\section{Highlighted Algorithms}

Utilizing the unified and extensible research framework provided by \easyvolcap, we incorporate state-of-the-art volumetric rendering algorithms and extract their reusable components as building blocks and inspirations for future research. Here we briefly introduce two of the most prominent ones.

\paragraph{Efficient neural radiance fields.} 
ENeRF \cite{lin2022efficient} is a generalizable neural radiance field method. 
Combining the power of cost-volume-based depth estimation and generalizability of image features (Image-Based Rendering), ENeRF achieves unprecedented performance and generalizability on both static and dynamic scenes. 
We extract the reusable MVSNeRF-style \cite{chen2021mvsnerf} cost-volume-based depth estimation module, IBRNet-style \cite{wang2021ibrnet} appearance learning and a depth-guided NeRF sampling pipeline like DONeRF \cite{neff2021donerf}.

\paragraph{Realtime rendering of dynamic volumetric video.} 
4K4D \cite{xu20234k4d} is a real-time 4D view synthesis method developed using the \easyvolcap framework.
We achieve 60FPS at 4K resolution on rendering of neural volumetric video by introducing hardware-accelerated differentiable depth-peeling algorithm on point clouds \cite{kerbl20233d,zhang2022differentiable}.
Through the parameter-sharing of the 4D structures, the geometry and appearance feature effectively fuse 4D information present in the multi-view video. 
Thanks to the compactness and expressiveness of the hybrid structure, we can pre-compute features on the explicit geometry during inference, achieving unprecedented rendering speed. 
The teaser and video of the brief are produced by this algorithm.

\paragraph{Other supported methods.} \easyvolcap also supports a wide range of other algorithms including \cite{muller2022instant,fridovich2023k,mildenhall2021nerf,park2021nerfies,wang2021neus,zhang2022differentiable,tancik2023nerfstudio} with much more coming.

\begin{table*}[t]
    \centering
    \vspace*{-1em}
    \caption{
        \textbf{Comparison of different viewers.}
        The web-based viewer adopted by NeRFStudio \cite{tancik2023nerfstudio} shows great extensibility, however, the unavoidable network transfer increases latency.
        Our native viewer excels in all these aspects thanks to the seamless integration of CUDA memory copy and the viewer's OpenGL context.
    }
    \vspace*{-1em}
    \resizebox{1.0\linewidth}{!}{
        \begin{tabular}{@{}ll*{4}{c}@{}}
            \toprule
            \textbf{Viewer}  & \textbf{Method}                                                   & \textbf{Low Latency} & \textbf{Async Drawing} & \textbf{Extensibility} & \textbf{Cross-Platform} \\
            \midrule
            WebViewer        & NeRFStudio\cite{tancik2023nerfstudio}                             & \xmark               & \xmark                 & \cmark                 & \cmark                  \\
            C++-based Viewer & InstantNGP\cite{muller2022instant}, 3D Gaussian\cite{kerbl20233d} & \cmark               & \cmark                 & \xmark                 & \cmark                  \\
            Python-based     & ENeRF\cite{lin2022efficient}, torch-ngp\cite{torchngp}            & \cmark               & \xmark                 & \cmark                 & \cmark                  \\
            \midrule
            Ours             & \easyvolcap                                                       & \cmark               & \cmark                 & \cmark                 & \cmark                  \\
            \bottomrule
        \end{tabular}
    }
    \label{tab:viewer_comparison}
\end{table*}

\begin{table*}[t]
    \centering
    \vspace*{-1em}
    \caption{
        \textbf{Comparison of the configuration system.}
        \textit{yacs} is the most common among open-source papers \cite{peng2022animatable}, however, it lacks an extensible and dynamic file type support, making it hard to scale.
        XRNeRF's \cite{xrnerf} register system shows great potential, but our configuration system handles inheritance between files better than theirs.
    }
    \vspace*{-1em}
    \resizebox{1.0\linewidth}{!}{
        \begin{tabular}{ll*{6}{c}}
            \toprule
            \textbf{Config System} & \textbf{Method}                        & \textbf{yaml} & \textbf{python} & \textbf{json} & \textbf{Command Line} & \textbf{Inheritance} & \textbf{Multi-Inheritance} \\
            \midrule
            yacs                   & ENeRF\cite{lin2022efficient}           & \cmark        & \xmark          & \xmark        & \cmark                & \xmark               & \xmark                     \\
            dataclasses            & NeRFStudio\cite{tancik2023nerfstudio}  & \xmark        & \cmark          & \xmark        & \cmark                & \cmark               & \xmark                     \\
            gin                    & MipNeRF-360\cite{barron2022mipnerf360} & \xmark        & \xmark          & \xmark        & \xmark                & \xmark               & \xmark                     \\
            register               & XRNeRF\cite{xrnerf}                    & \cmark        & \cmark          & \cmark        & \cmark                & \cmark               & \xmark                     \\
            \midrule
            Ours                   & \easyvolcap                            & \cmark        & \cmark          & \cmark        & \cmark                & \cmark               & \cmark                     \\
            \bottomrule
        \end{tabular}
    }
    \label{tab:config_comparison}
    \vspace*{-1em}
\end{table*}
\section{Framework Utilities}

\subsection{High-Performance Native Viewer}

\easyvolcap provides a high-performance native viewer that delivers the rendered content of various custom algorithms to the user's screen with low latency, high throughput, and extensibility by incorporating the easy-to-use Python language to define control elements.
A comparison with other common open-source implementations can be found in Table~\ref{tab:viewer_comparison}.

\paragraph{CPU-GPU communication.} 
\easyvolcap's viewer implementation fully harnesses the power of asynchronous CUDA kernel launching to overlap the slow Python-based user interface with GPU-side network rendering. 
Moreover, by directly utilizing the CUDA-Graphics interface, \easyvolcap copies the rendered content stored in a PyTorch tensor directly to the frame buffer of the OpenGL context to be displayed on the screen. 
This design and implementation choice avoids a heavy GPU-CPU-GPU copy chain and their respective synchronization points, greatly enhancing the throughput and reducing the latency and overhead of the viewer.

\paragraph{Memory Management.} 
Although a single frame being rendered is usually small enough to be stored directly in the VRAM, most dynamic volumetric video representations are too large to directly fit onto the VRAM of the Graphics Card. 
Taking inspiration from other video playback software \cite{potplayer}, \easyvolcap heuristically swaps out the VRAM with the immediately following frames to be played on the pinned main memory with asynchronous and non-blocking copy \cite{memory} to satisfy both the computing and PCIe hardware, achieving maximum rendering speed.

\subsection{Logistics Systems}

\paragraph{Robust configuration system.} 
\easyvolcap extends the configuration system of XRNeRF \cite{xrnerf} by incorporating their registration system and extending their yaml-based configuration system with the support of the command line tab-complete.
Inheritance from multiple different config files is also supported to plug in various different settings conveniently, along with an extended reference system that supports constructing configurations using the dynamic values in other config files.
Such adaptation allows for an easy plug-and-play experience to swap out new datasets, entirely new algorithms, or just replace a hyperparameter setting directly from the command line to perform new experiments. 
A comparison with other popular configuration frameworks is provided in Table~\ref{tab:config_comparison}.

\begin{acks}
    The authors would like to acknowledge support from NSFC (No. 62172364), Information Technology Center, and State Key Lab of CAD\&CG, Zhejiang University.
\end{acks}

\bibliographystyle{ACM-Reference-Format}
\bibliography{reference}
\appendix

\end{document}